\documentclass{article}

\usepackage{microtype}
\usepackage{amssymb} 
\usepackage{graphicx}
\usepackage{bbding}
\usepackage{subcaption}
\usepackage{booktabs} 
\usepackage{longtable}
\usepackage{multirow}
\usepackage{mathrsfs}
\usepackage{booktabs} 
\usepackage{breqn} 
\usepackage{float}
\usepackage{hyperref}

\usepackage[accepted]{icml2024}

\usepackage{amsmath}
\usepackage{amssymb}
\usepackage{mathtools}
\usepackage{amsthm}

\usepackage[capitalize,noabbrev]{cleveref}

\theoremstyle{plain}

\theoremstyle{definition}

\theoremstyle{remark}

\usepackage[textsize=tiny]{todonotes}

\icmltitlerunning{DGCformer}

\begin{document}

\twocolumn[
\icmltitle{DGCformer: Deep Graph Clustering Transformer for Multivariate \\Time Series Forecasting}



\icmlsetsymbol{equal}{*}
\icmlsetsymbol{co}{$\dag$}

\begin{icmlauthorlist}
\icmlauthor{Qinshuo Liu}{1}
\icmlauthor{Yanwen Fang}{1}
\icmlauthor{Pengtao Jiang}{2}
\icmlauthor{Guodong Li}{co,1}
\end{icmlauthorlist}

\icmlaffiliation{1}{Department of Statistics and Actuarial Science, The University of Hong Kong, Hong Kong SAR, China}
\icmlaffiliation{2}{Vivo Mobile Communication Co., Ltd.  Hangzhou China}

\icmlcorrespondingauthor{Guodong Li}{gdli@hku.hk}

\icmlkeywords{Time Series Forecasting, Clustering, Graph network, Deep Learning}
\vskip 0.3in
]



\printAffiliationsAndNotice{} 

\begin{abstract}
Multivariate time series forecasting tasks are usually conducted in a channel-dependent (CD) way since it can incorporate more variable-relevant information. However, it may also involve a lot of irrelevant variables, and this even leads to worse performance than the channel-independent (CI) strategy.
This paper combines the strengths of both strategies and proposes the Deep Graph Clustering Transformer (DGCformer) for multivariate time series forecasting.
Specifically, it first groups these relevant variables by a graph convolutional network integrated with an autoencoder, and a former-latter masked self-attention mechanism is then considered with the CD strategy being applied to each group of variables while the CI one for different groups.
Extensive experimental results on eight datasets demonstrate the superiority of our method against state-of-the-art models, and our code will be publicly available upon acceptance.

%
%
\end{abstract}

\section{Introduction}

Multivariate Time Series Forecasting (MTSF) has gained popularity in various domains, including finance \citep{patton2013copula}, electricity \citep{amarawickrama2008electricity}, energy \citep{demirel2012forecasting,deb2017review}, etc.
%
It has been greatly improved and brought to a new stage by using deep neural networks, such as CNN and RNN; see \citet{hochreiter1997long,chung2014empirical,li2017diffusion,liu2020dstp,qin2017dual}.
In the meanwhile, many transformer-based methods \cite{vaswani2017attention} have also been employed to learn long-range temporal dependency in multivariate time series \cite{li2019enhancing,zhou2021informer,wu2021autoformer,zhou2022fedformer,liu2021pyraformer,zhang2022crossformer,nie2022time,ekambaram2023tsmixer}.




For a MTSF task, it is important to explore the interrelationships among various channels or variables, and the predictive accuracy for a certain variable can be improved by incorporating more relevant information from other variables.
For example, in forecasting air temperature, it is beneficial to consider historical humidity data, as it provides pertinent contextual information that can improve the prediction precision. 
Several models have been proposed to capture these interrelationships utilizing CNN, RNN and/or graph convolution layers; see, e.g., the LSTNet in \citet{lai2018modeling} and MTGNN in \citet{wu2020Connecting}. 
Moreover, Crossformer \cite{zhang2022crossformer} applied the attention mechanism to both the time and variable (channel) dimensions, and TSmixer \citep{ekambaram2023tsmixer} utilizes time-mixed and feature-mixed MLP for aggregating information. 
All the above methods adopt the so-called channel-dependent (CD) modeling strategy.


\begin{figure}[t]
\setlength\tabcolsep{1pt}
    \centering
    \includegraphics[width=0.47\textwidth]{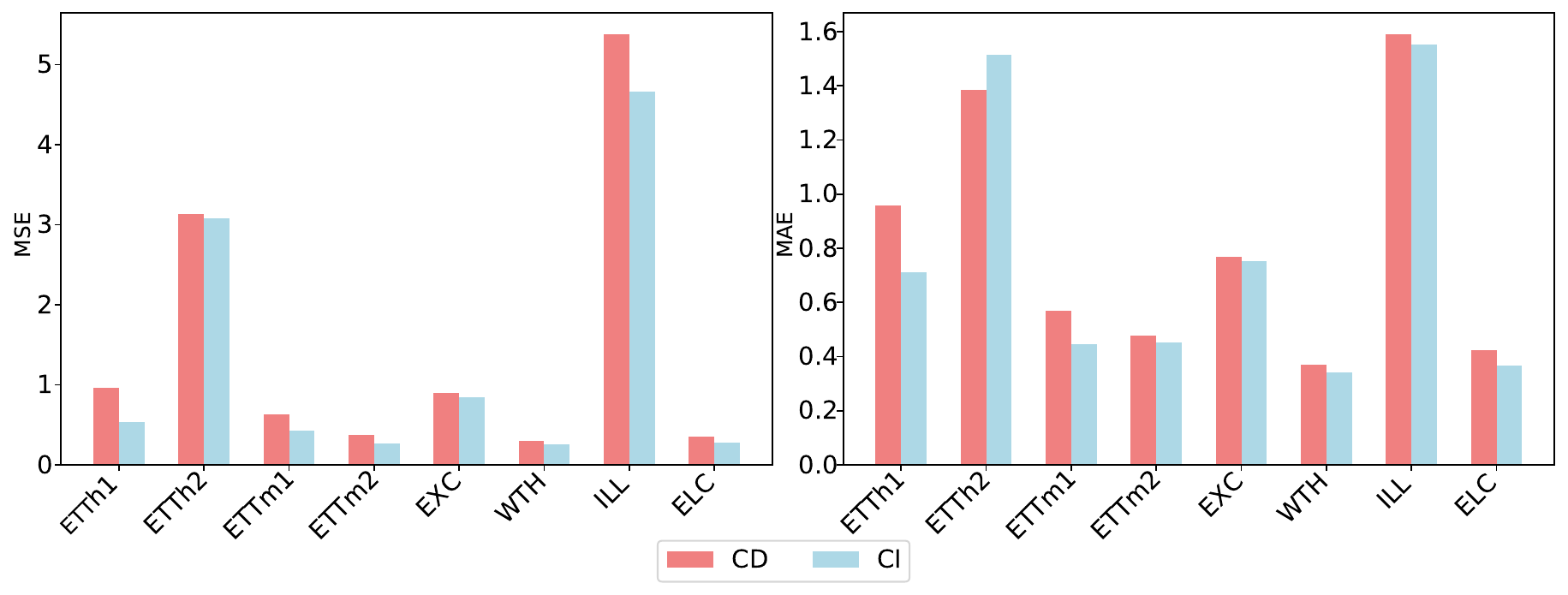}
    \caption{MSEs and MAEs of Informer trained on eight common datasets with the CI and CD strategies.}
    \label{fig:barcicd}
\end{figure}


Although the CD strategy can lead to superior results across a majority of datasets, it may lead to performance drop due to the fact that it takes into account some interference signals from irrelevant variables. 
Considering these factors, CI would be a more reasonable choice to model the features separately, and this is supported by the empirical evidences in  \citet{han2023capacity} and Figure \ref{fig:barcicd}.
\citet{han2023capacity} utilized the autocorrelation functions (ACF) to measure distribution drift which is one of the factors leading to inferior performances of CD methods.
They found that the values of ACF for training and testing sets varied significantly across various datasets, which means the existence of distribution drift in these datasets.

\begin{figure}[htbp]
    \centering
    \includegraphics[width=0.45\textwidth]{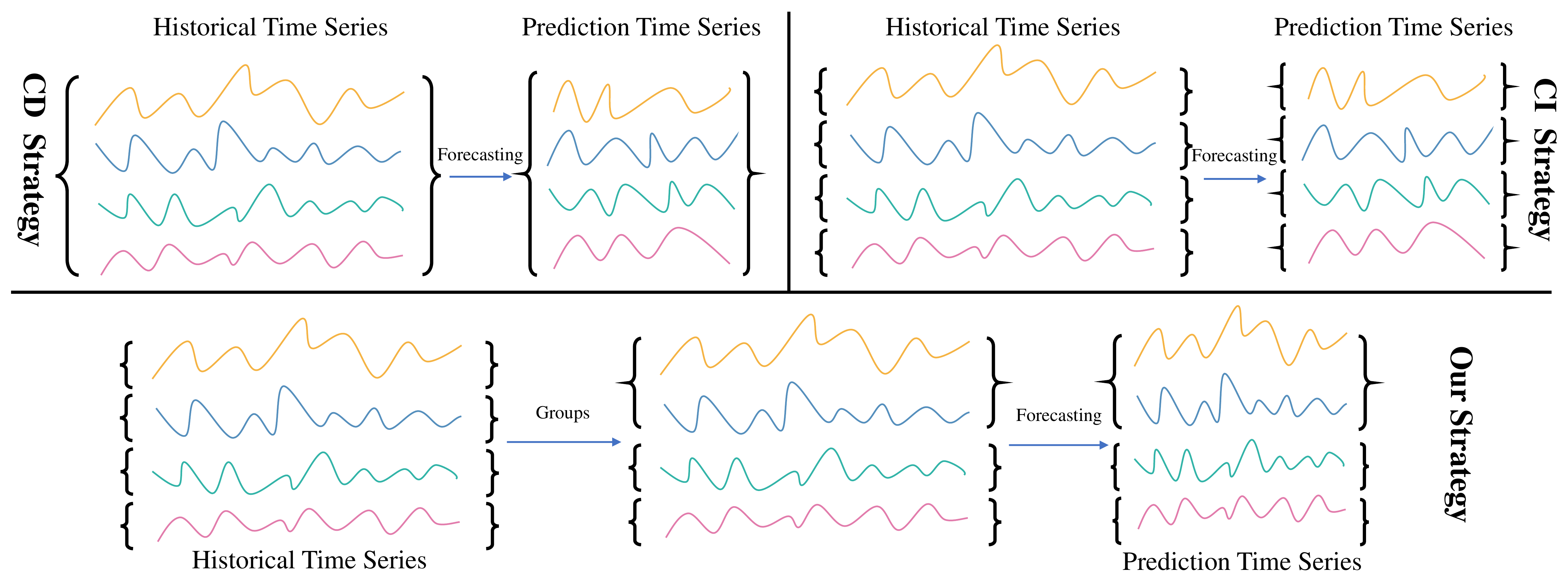}
    \caption{Comparison of the CI and CD strategies, as well as our DGCformer. At the upper panel, the CD strategy is depicted at the left, where each channel relies on the historical data of all channels, while the CI strategy is illustrated at the right, featuring channels that operate independently. 
    Our method is showcased at the lower panel, learning the inner-cluster channel-wise dependencies.}
    \label{fig:CICD}
\end{figure}

Since both strategies have their own strengths, then it is natural to combine them by proposing a new method for MTSF.
The resulting model is called Deep Graph Clustering Transformer (DGCformer), whose main idea is illustrated in Figure \ref{fig:CICD}.
Specifically, we first form clusters for variables with significant similarities and treat them as one group, while channels with low similarities are allocated to distinct groups. 
The CD strategy is then applied to the variables within each cluster, while the CI one is employed for the variables among different clusters.
As a result, the new strategy can improve forecasting accuracy and interpretability\footnote{In this paper, we utilize the concept of interpretability to elucidate the natural interdependence that exists among variables in real-world data. For instance, the relationship between wind and temperature needs to be taken into consideration as they influence each other.}.

%
%
%

The proposed DGCformer has two stages: \textbf{Deep Graph Clustering} and \textbf{Transformer Forecasting}. The clustering stage draws inspiration from \citet{bo2020structural}, and it integrates graph convolutional network (GCN) \citep{defferrard2016convolutional} and autoencoder-based layers \citep{kingma2013auto} to supervise  channels classification. 
Then, a former-latter masked self-attention mechanism is proposed, which computes the temporal similarity and the pairwise similarity among channels within the groups. 
Notably, our approach emphasizes relationships solely within the same cluster during the grouped variable-dimension step. 
This method effectively merges information from the data source, and it hence accurately captures distinct group characteristics. Our main contributions are four-fold:

\begin{itemize}
\item We present DGCformer, a novel transformer-based model, designed to significantly enhance predictive capabilities in the MTSF domain. This innovation confirms the utility of harnessing the combined strengths inherent in both CI and CD strategies.

\item By identifying and considering the substantial similarities among variables, we have developed the Deep Graph Clustering module. This module, grounded in GCN and autoencoder layers, adeptly consolidates channels into discrete clusters.

\item We organize the clustered series into segments and deploy a former-latter masked self-attention mechanism that discerns the interconnections within each cluster following the CD strategy. Conversely, for channels segmented into disparate groups, we apply the CI approach to maintain the independency.



\item The feasibility of our proposed method is evaluated by comparing ours with other sota models on eight datasets. 
Remarkably, our approach exhibits superior performance, ranking top-1 among the 7 models in the 57 comparisons out of 64 ones with various prediction lengths and evaluation metrics. 
\end{itemize}

\section{Related Work}


\textbf{Preliminary neural networks for time series forecasting.} In the realm of long-term time series forecasting, considerable efforts have been directed towards employing computer science-based methodologies. Before the widespread integration of transformer-based models in this domain, Convolutional Neural Network (CNN) and Recurrent Neural Network (RNN) are predominantly used for such analyses. CNN plays a pivotal role in extracting information on temporal variations, as delineated in \citet{hochreiter1997long,rangapuram2018deep}. Concurrently, RNN predicts future data points adhering to the Markov assumption, a concept further explored in \citet{lecun1998gradient}.
Recent advancements in RNN technology have led to the development of innovative approaches like Long Short-Term Memory (LSTM) \cite{hochreiter1997long} and Gated Recurrent Unit (GRU) \cite{chung2014empirical} networks. These advanced networks incorporate gating mechanisms that effectively control information flow, thereby significantly outperforming traditional RNN architectures in terms of performance.

\textbf{Transformers for MTSF.} In the context of MTSF, many transformer-based techniques have appeared. LogTrans \cite{li2019enhancing} integrates convolutional self-attention layers with a LogSparse structure, facilitating local information capture while diminishing space complexity. Further advancements introduce embeddings that incorporate scalar projections, as well as local and global time stamps \cite{li2019enhancing}. Informer \cite{zhou2021informer} refines the attention mechanism by concentrating on the top-k significant keys, effectively reducing time complexity from $O(N^2)$ to $O(N\log N)$, a notable improvement over conventional transformer models. Autoformer \cite{wu2021autoformer} innovates by decomposing time series into seasonal and trend components. Pyraformer \cite{liu2021pyraformer} utilizes a pyramidal attention structure with both inter-scale and intra-scale connections to also maintain linear complexity. FEDformer \cite{zhou2022fedformer} employs a fourier-enhanced framework to transition series from time to frequency domain, thereby achieving linear complexity. Crossformer \cite{zhang2022crossformer} investigates the interrelations among multiple time series. Additionally, PatchTST \cite{nie2022time} introduces a channel independence system and a patching method, which we adopt as the baseline model for our analysis.


\textbf{MTSF with CD strategy.} Recently, there are some approaches explored to capture correlations between different variables. Statistical models like the Bayesian Structural Time Series model \citep{qiu2018multivariate} offers flexible modeling of dependencies and correlations in multivariate time series. Neural models, such as Deep State Space Models (DSSMs) \citep{rangapuram2018deep}, Temporal LSTnet \citep{lai2018modeling}, and Convolutional Network (TCN) \citep{wan2019multivariate}, aim to capture variable correlations in MTSF. Additionally, research has focused on utilizing graph neural networks (GNNs) \citep{wu2020Connecting} to explicitly capture cross-dimension dependency structures. Recent models like Crossformer \citep{zhang2022crossformer} and TSmixer \citep{ekambaram2023tsmixer} have proposed using transformers and MLPs to efficiently capture correlations among variables.

\textbf{Clustering methods.} In the context of clustering, traditional methods like the k-means algorithm rely on a limited set of features extracted from time series data, such as mean and variance, and consider the Euclidean distance between two series. However, this approach may not effectively capture the complexities associated with time series data. To address this challenge, novel strategies have been proposed. For example, \citet{paparrizos2015k} introduces the k-shape clustering algorithm, which employs the Dynamic Time Warping (DTW) \citep{muller2007dynamic} distance to capture temporal dependencies and variations within time series data more effectively. Additionally, methods based on Variational Autoencoders \citep{dilokthanakul2016deep} have been developed to derive more accurate cluster features based on the original data.


\section{Method}
\subsection{Preliminary}
\label{sec:preli}
In MTSF, given historical observations, denoted as $\mathbf{X} = \{\mathbf{x}_1,\mathbf{x}_2,\cdots,\mathbf{x}_N\} \in \mathbb{R}^{N \times T}$ with $N$ variables and $T$ time steps.  Then, the objective is to predict the future value $S$ time steps $\mathbf{Y} = \{\mathbf{y}_1,\mathbf{y}_2,\cdots,\mathbf{y}_N\} \in \mathbb{R}^{N \times S}$.

\begin{figure}[t]
\centering
\begin{subfigure}{0.235\textwidth}
\begin{center}
\includegraphics[width = 1.0\textwidth]{"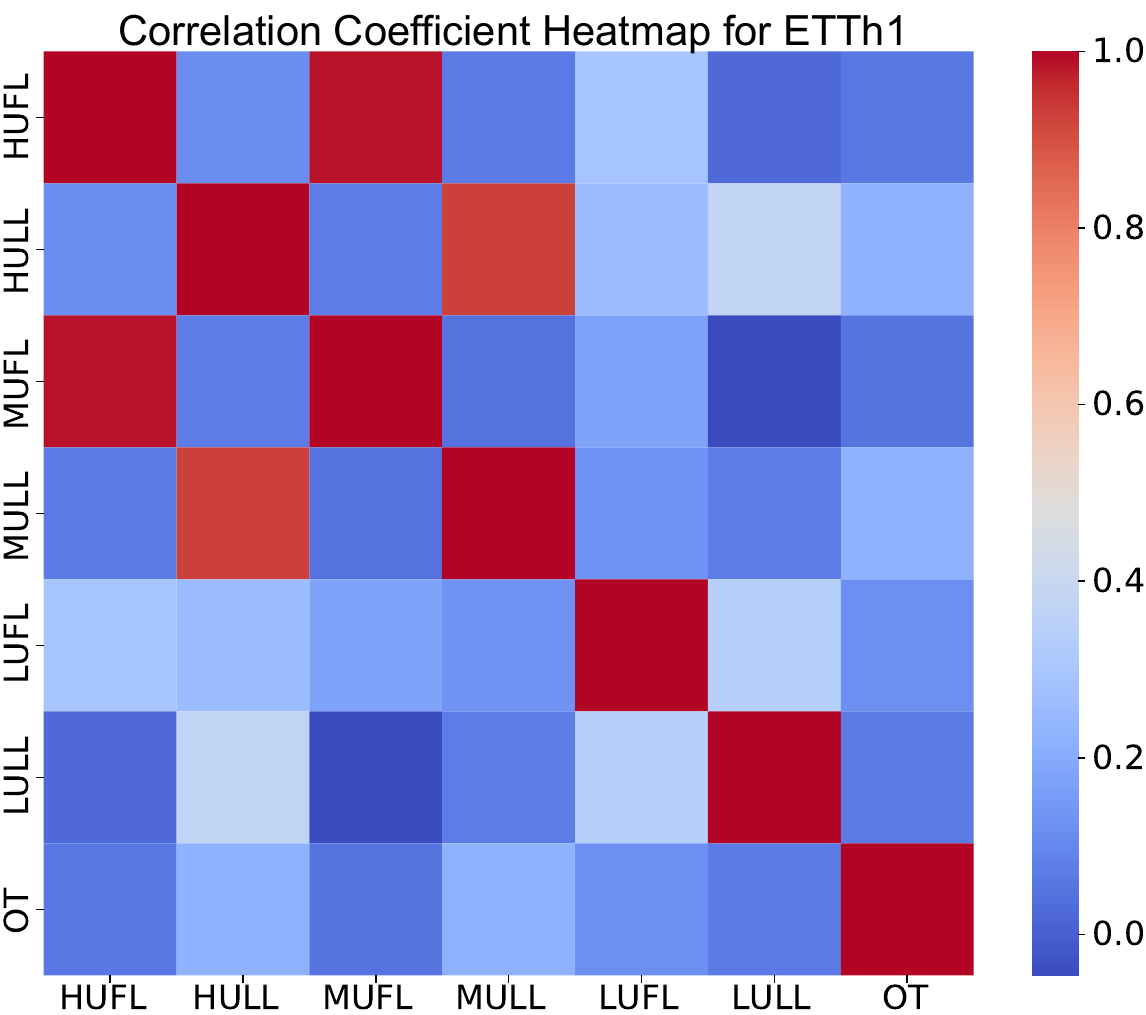"}
\caption{}
\end{center}
\end{subfigure}
\begin{subfigure}{0.235\textwidth}
    \begin{center}
        \includegraphics[width = 1.0\textwidth]{"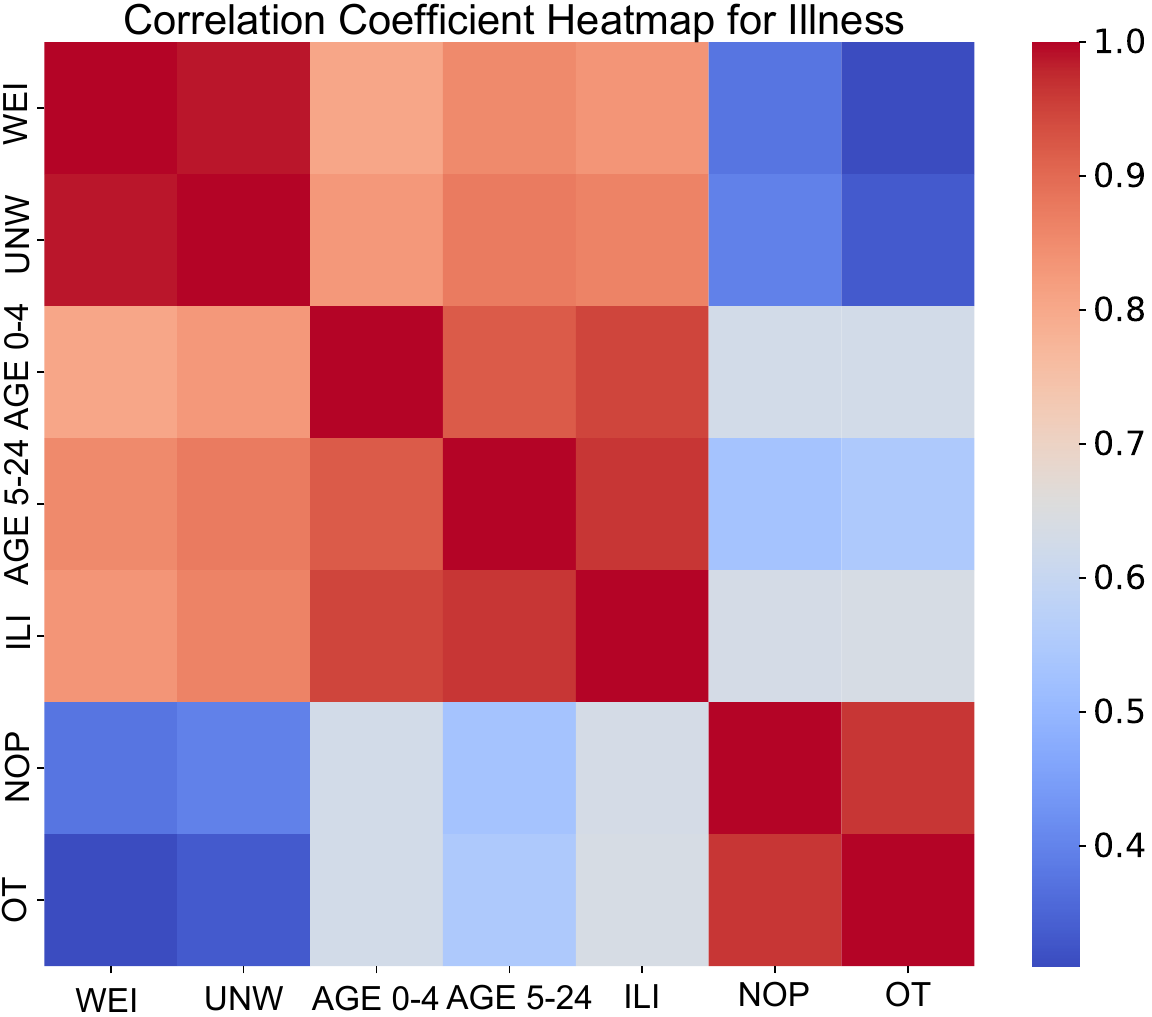"}
        \caption{}
    \end{center}
\end{subfigure}
\caption{
Figure(a) and (b) depict the correlations of channels in the ETTh1 and Illness datasets, respectively.}
\label{fig:heatmap}
\end{figure}

Before presenting our approach, we conduct an analysis of the inter-variable correlation within specific time series datasets. Figure \ref{fig:heatmap}(a) and (b) depict the heatmaps derived from the ETTh1 and Illness datasets, respectively. These two heatmaps reveal notable associations among certain variables, while also uncover the absence of significant similarities among others. Such findings serve as a compelling motivation for our endeavor to group these variables, thereby facilitating more accurate forecasting.
\begin{figure*}[htbp]
    \centering
    \includegraphics[width = 1.0\textwidth]{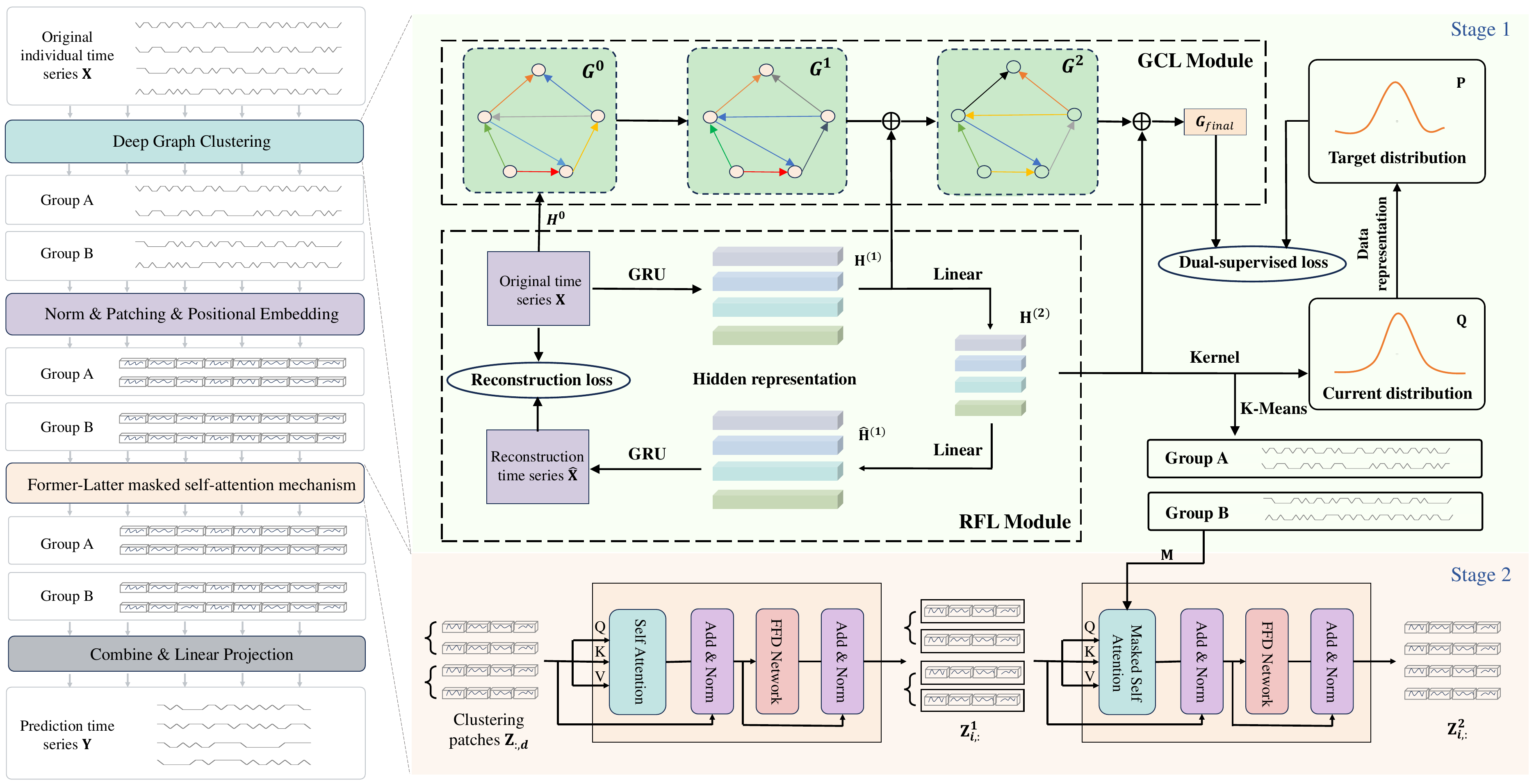}
    \caption{The schematic delineates the DGCformer method's workflow. The left section offers a comprehensive view of the entire pipeline. The right section focuses on the intricate deep graph clustering process, which groups channels into clusters, and the next former-latter masked self-attention mechanism that discerns relationships between channels within groups. In the implementation of the former-latter masked self-attention mechanism, the variables denoted by $\mathbf{Z}^1_{i,:}$ are processed as input tokens for the subsequent stage of the attention mechanism, which is delineated by black boxes.}
    \label{fig:overview}
\end{figure*}

Hence, we propose a novel approach termed DGCformer. The schematic overview of our DGCformer model is depicted in Figure \ref{fig:overview}. To leverage variable correlations while ensuring robustness, in Section \ref{sec:cluster}, we introduce a deep graph clustering method to group the variables. More specifically, within the deep clustering phase, we integrate a graph convolutional network with an autoencoder architecture to capture low and high-order structural information, representing direct and indirect relationships between variables. In Section \ref{sec:forecast}, we partition the grouped series into patches and introduce a former-latter masked self-attention mechanism to capture the time-variable-dimension similarity within groups. After this, a simple linear layer is employed for final prediction.

\subsection{Deep Graph Clustering}
\label{sec:cluster}

In this section, we propose the methodology employed for segregating channels into distinct clusters. The aim of clustering is to divide $\mathbf{X}$ into $n$ groups as $\{\mathbf{X}^{(1)},\mathbf{X}^{(2)},\cdots,\mathbf{X}^{(n)}\}$, where each group $\mathbf{X}^{(i)}$ is fed independently into transformer structure. Our strategy encompasses two modules: the reconstruction-feature learning module (RFL) and the graph-cluster learning (GCL) module. The primary objective of the RFL module is to extract a feasible low-dimensional representation, which is conducive to effective clustering. Here, we have opted for a straightforward AE for its generality and a GRU network to supplant the traditional linear model, which is instrumental in capturing temporal features. Conversely, the purpose of GCL is to outline the direct relationships among variables. Given the popularity and efficacy of GCNs in capturing such inter-variable relationships, it was a natural choice for this role. The GCL mechanism is adept at learning dual facets of information: the intrinsic properties of the variables and the interconnections between them. To preserve and leverage the information encapsulated in each AE layer, we integrate the representations derived from both the GCN and GRU networks within the GCL and RFL submodules. Our method is achieved by discerning similarities within the variables and computing the probability of these variables being associated with the user-defined cluster centers. This dual-module approach facilitates a comprehensive understanding of the data's relationship between channels.

\textbf{Reconstruction-feature learning module.} At the core of this module lies the use of the autoencoder technique, a staple in data analysis, to distill low-dimensional representations from the raw data. These representations are crucial for clustering and the effectiveness of these is instrumental in facilitating thorough and insightful analysis.

The encoder's architecture features a Gated Recurrent Unit (GRU) layer as the primary information extractor, crucial for capturing the temporal dynamics inherent in time series data. Following this, a fully connected layer transforms the data into the desired latent variable representation, $\mathbf{H}^{(2)}$. This efficient representation is accomplished using just two layers.
\begin{equation}
\begin{aligned}
    \mathbf{H}^{(1)} & =\phi (\texttt{GRU}(\mathbf{X})),\\
    \mathbf{H}^{(2)} & =\phi (\mathbf{W}^{enc} \mathbf{H}^{(1)}+\mathbf{b}^{enc}),
\end{aligned}
\end{equation}
where $\phi$ signifies the activation function, commonly selected as the Rectified Linear Unit (ReLU). The weight matrix and bias are denoted by $\mathbf{W}^{enc} \in \mathbb{R}^{l_2 \times l_1}$ and $\mathbf{b}^{enc} \in \mathbb{R}^{l_2}$, respectively. Here, $l_2$ represents the latent dimension post-linear transformation, and $l_1$ indicates the latent dimension following the GRU network.

Transitioning to the decoder network, the latent variable is employed as the starting point for the reconstruction of the input time series. This reconstruction process involves a fully connected layer succeeded by a GRU layer. Echoing the encoder's framework, the output of the decoder layer, denoted as $ \widetilde { \mathbf { X } } $, is formulated through analogous equations:
\begin{equation}
\begin{aligned}
    & \widetilde{\mathbf{H}}^{(1)} =\phi (\mathbf{W}^{dec} \mathbf{H}^{(2)}+ \mathbf{b}^{dec}),\\
    & \widetilde{\mathbf{X}}  =\phi (\texttt{GRU}(\widetilde{\mathbf{H}}^{(1)})),
\end{aligned}
\end{equation}
where $\mathbf{W}^{dec} \in \mathbb{R}^{l_1 \times l_2}$ and $\mathbf{b}^{dec} \in \mathbb{R}^{l_1}$ represent the weight matrix and bias for decoder structure.

Upon acquiring the latent representation from the autoencoder framework, a t-distribution \citep{van2008visualizing} is employed to evaluate the similarity between the embedded node $\mathbf{H}^{(2)}$ and the $n$ cluster center nodes, denoted as $\boldsymbol{\mu}$. This distribution of similarity is conceptualized as the probability of node $i$ being part of cluster $j$. The expression for this probability is as follows:
\begin{equation}
q_{i j}=\frac{\left(1+\left\|h_i-\mu_j\right\|^2 / t\right)^{-\frac{t+1}{2}}}{\sum_{j^{\prime}}\left(1+\left\|h_i-\mu_{j^{\prime}}\right\|^2 / t\right)^{-\frac{t+1}{2}}},
\end{equation}
where $h_i$ represents the $i$-th row of the embedded node matrix $\mathbf H^{(2)}$, and $\mu_j\in \boldsymbol{\mu}$ denotes the initialization learned by the pre-trained autoencoder. The parameter $t$ is determined by our selection, and usually set by $t=1$. After obtaining the distribution $q_{ij} \in \mathbf{Q}$, we calculate the target distribution $p_{ij} \in \mathbf{P}$ aiming to make data representation closer to cluster centers which can be expressed as follows:
\begin{equation}
p_{i j}=\frac{q_{i j}^2 / f_j}{\sum_{j^{\prime}} q_{i j^{\prime}}^2 / f_{j^{\prime}}},
\end{equation}
where $f_j = \sum _i q_{ij}$. This target distribution, denoted as $ \mathbf { P } $, is utilized to supervise the distribution in the following module.

\textbf{Graph-cluster learning module.} The GCL module initiates with the construction of a graph to delineate direct relationships among variables. This is followed by the incorporation of a GCN module. This module is pivotal in both assimilating information within the variables and estimating the probability of these variables being associated with the provided cluster centers.

The process begins with the formation of an initial directed graph, denoted as $ \mathbf { G}^{(0)} $, utilizing our original series $ \mathbf { X } $. We initialize its adjacency matrix, $ \mathbf { A } $, using the correlation matrix, with a threshold defined by the user.

Moreover, to retain the multi-dimensional information yielded from the outputs of the GRU and Linear layers, the representations extracted from each layer are amalgamated into the respective GCN layer. This integration is crucial for the efficient dissemination of information across the network, thereby enhancing the propagation and utilization of these learned representations. In the terminal stage of the GCN, a softmax layer is implemented to get the output, $ g_{ij} \in \mathbf{G}_{final} $, which also signifies the probability of variable $ i $ being associated with clustering center $ j $. This output is interpreted as a distribution for comparison against the guiding distribution derived from the latent variable in the RFL module. The formula of GCN is shown as follows:
\begin{equation}
\begin{aligned}
    &\mathbf G^{(1)} =\phi\left(\widetilde{\mathbf D}^{-\frac{1}{2}} \widetilde{\mathbf A} \widetilde{\mathbf D}^{-\frac{1}{2}} \mathbf X \mathbf W^{(0)}\right),\\
        &\widetilde{\mathbf G}^{(1)} = (1-\epsilon) {\mathbf G}^{(1)} + \epsilon \mathbf H^{(1)},\\
        &\mathbf G^{(2)}  =\phi\left(\widetilde{\mathbf D}^{-\frac{1}{2}} \widetilde{\mathbf A} \widetilde{\mathbf D}^{-\frac{1}{2}} \widetilde{\mathbf G}^{(1)} \mathbf W^{(1)}\right), \\
         &\widetilde{\mathbf G}^{(2)} = (1-\epsilon) {\mathbf G}^{(2)} + \epsilon \mathbf H^{(2)},\\
        & \mathbf{G}_{final} =\texttt{softmax} \left(\widetilde{\mathbf{D}}^{-\frac{1}{2}} \widetilde{\mathbf{A}} \widetilde{\mathbf D}^{-\frac{1}{2}} \widetilde{\mathbf G}^{(2)} \mathbf W^{(2)}\right), \\
\end{aligned}
\label{GCN equation}
\end{equation}
where
$$
\widetilde{\mathbf A}=\mathbf A+\mathbf I \text { and } \widetilde{\mathbf D}_{i i}=\sum_j \widetilde{\mathbf A}_{i j}.
$$
Here $\mathbf{W}^{(0)} \in \mathbb{R}^{L \times l_1}$, $\mathbf{W}^{(1)} \in \mathbb{R}^{l_1 \times l_2}$, and $\mathbf{W}^{(2)} \in \mathbb{R}^{l_2 \times n}$ represent weight matrices, where $L$ denotes the length of the series and $n$ indicates the number of clusters as specified in \ref{sec:cluster}. For the hyperparameter $\epsilon$, we adhere to the standard value of 0.5.

\textbf{K-means Cluster.} 
In this phase of our analysis, we employ a straightforward K-means clustering method to categorize the hidden layer $\mathbf H^{(2)}$ derived from the RFL module. Given that $\mathbf H^{(2)}$ effectively represents the original time series, clustering it can be considered equivalent to clustering the original series itself. This approach offers the advantage of reducing the dimensionality of the time series length, thereby conserving computational time. Furthermore, the outcomes of this clustering process also named cluster results are utilized as the basis for generating the mask matrix, a crucial component in our masked transformer forecasting module.

\textbf{Clustering Loss.} The clustering loss function in our model is divided into two components: the reconstruction loss and the dual-supervised loss.

By juxtaposing the reconstructed series with the original series, we evaluate the latent series' effectiveness in the clustering context. This comparison offers insights into the latent series’ ability to discern and encapsulate significant clusters within the data. The reconstruction loss is quantified using the Mean Squared Error (MSE) loss function, serving as the primary metric for this evaluation:
\begin{equation}
    \mathcal{L}_{\texttt{REC}} = \frac{1}{2N}\sum_{i=1}^N \|\widetilde{\mathbf{X}}-\mathbf{X}\|_F^2.
\label{loss rec}
\end{equation}
For the dual-supervised loss, we employ the target distribution delineated in the construction module to guide the distribution $g_{ij} \in \mathbf{G}_{final}$, utilizing the Kullback-Leibler (KL) divergence. The loss function specific to the dual-supervised is defined as follows:
\begin{equation}
\mathcal{L}_{\texttt{DS}}=K L(\mathbf P \| \mathbf G_{final})=\sum_{i=1}^N \sum_{j=1}^{n} p_{i j} \log \frac{p_{i j}}{g_{i j}}.
\label{loss GCN}
\end{equation}

\subsection{Masked Transformer Forecasting}
\label{sec:forecast}

In this section, our method builds on the foundational methodology introduced in PatchTST \cite{nie2022time}. We propose a novel former-latter masked self-attention mechanism module to capture the dependency information within grouped variables which to substantially enhance its performance 

This forecasting process relies on the inherent correlation between any two time series. To facilitate this, we utilize the clustering results to generate a binary cluster-based attention mask matrix $\mathbf M \in \mathbb{R}^{N \times N}$, where $\mathbf M_{ij}=1/0$ indicates that variables $i$ and $j$ belong to/not belong to the same cluster. Once we have obtained the $n$ groups $\{\mathbf{X}^{(1)},\mathbf{X}^{(2)},\cdots,\mathbf{X}^{(n)}\}$, we follow the next to process the variables grouped:

\textbf{Normalization and Patching:} To effectively handle the time series data, we first normalize and segment the series into patches, a method we refer to as ``norm \& patching''. For each univariate time series $\mathbf{x}_i$, where $i\in {1, \ldots, N}$, we introduce a constant parameter $C$ to represent the desired number of patches. This parameter choice is in line with the approach detailed in \citet{nie2022time}. As a result, the dimensionality of each time series expands to $C \times d$, with $d$ being the length of each patch. After this we introduce a time-variable-dimension attention block to extract pertinent information.

\textbf{Former-latter masked self-attention mechanism:} In this part, we propose an attention module that focuses on extracting relevant information from the series. The objective is to identify correlations between any two patches within a series and to investigate interactions within any group $\mathbf{X}^{(i)},i=1, \ldots, n$.

This module first includes a multi-head attention block, an add \& norm block, and a forward feed network (FFN) with add \& norm. This structure yields a new representation for each time series, highlighting correlations between patches. We then introduce a mask self-attention block to capture relationships between variables in every cluster $\mathbf{X}^{(i)},i=1, \ldots, n$ given in Section \ref{sec:cluster}. Utilizing self-attention across all $N$ series, the mask dictates the computation of similarities with other variables. Additionally, we integrate a mask vector, $M \in \mathbb{R}^{N \times 1}$, to govern the inclusion of FFN and dropout layers following the attention block. The mask vector $M$ is defined as follows:
\begin{equation}
 M_i= M_j = 
\left\{
\begin{array}{cc}
 1 & \text{if} \quad (\mathbf M_{ij} \neq 0 ),\\
 0 & \text{otherwise}.
\end{array}
\right.
\end{equation}
The mathematical expressions for two attention blocks are as follows:
\begin{equation}
	\begin{aligned}
	&	\mathbf {\hat Z}_{:,d}  = \texttt{LayerNorm}(\mathbf Z_{:,d} + \texttt{MSA}(\mathbf Z_{:,d},\mathbf Z_{:,d},\mathbf Z_{:,d})), \\
	&	\mathbf Z^1  = \texttt{LayerNorm}(\mathbf {\hat Z}_{:,d} + \texttt{MLP}(\mathbf {\hat Z}_{:,d})), \\
	&	\mathbf {\widetilde Z}_{i,:} = \texttt{LayerNorm}(\mathbf Z^1_{i,:} + \texttt{MSA}(\mathbf Z^1_{i,:},\mathbf Z^1_{i,:},\mathbf Z^1_{i,:},\mathbf{M})), \\
	&	\mathbf Z^2 = \texttt{LayerNorm}(\mathbf {\widetilde Z}_{i,:} + \texttt{MLP}(\mathbf {\hat Z}_{i,:})).
	\end{aligned}
\end{equation}
Here, $\mathbf{Z}^1 \in \mathbb{R}^{N \times C \times d}$ is the self-attention output, and $\mathbf{Z}^2 \in \mathbb{R}^{N \times C \times d}$ represents the masked self-attention output, enabling interactions within groups via mask matrix $\mathbf{M}$. $\texttt{MSA}$ denotes multi-head self-attention, $\texttt{MLP}$ refers to the multi-layer feedforward network, and $\texttt{LayerNorm}$ signifies layer normalization, essential in transformer architectures \cite{dosovitskiy2020image}.

Subsequent to the former-latter masked self-attention mechanism, a linear projection \citep{zeng2023transformers} is applied to generate predictions. The dimension of the resulting prediction, $\widehat{\mathbf{Y}} \in \mathbb{R}^{N \times S}$ is same as $\mathbf{Y}$, shown in Section \ref{sec:preli}.

\subsection{Optimization Objective}

The total loss function is composed of three distinct elements: the reconstruction loss, the dual-supervised loss, and the prediction loss, denoted as $\mathcal{L}_{\texttt{PRED}}$. As a result, the aggregate form of the loss function is mathematically articulated, incorporating coefficients $\lambda_1=0.1$ and $\lambda_2=1$, as follows:
\begin{equation}
\label{loss_func}
    \mathcal{L}_{\texttt{total}} = \lambda_1 \mathcal{L}_{\texttt{DS}} + \lambda_2 \mathcal{L}_{\texttt{REC}} + \mathcal{L}_{\texttt{PRED}}, 
\end{equation}
where
\begin{equation}
    \mathcal{L}_{\texttt{PRED}} = \frac{1}{2N}\sum_{i=1}^N \|\widehat{\mathbf{Y}}-\mathbf{Y}\|_F^2.
\end{equation}

\section{Experiments}
\subsection{Dataset and Experiment Details}
\label{dataset}
\textbf{Dataset.} Our experiments utilize a diverse array of datasets, commonly employed in the MTSF problems. We have conducted comprehensive experiments on eight distinct datasets, namely: ETTh1, ETTh2, ETTm1, ETTm2, Weather, Illness, Electricity, and Exchange. A brief overview of each dataset is shown in Appendix \ref{app:Dataset}.

\begin{table*}[t]
\caption{Full results for the long-term forecasting task are presented in this section. The input sequence length is set to 104 for the ILL task, while it is 96 for the remaining tasks. For the ILL dataset, we evaluate prediction lengths ($S$) of 24, 36, 48, and 60, whereas for the other datasets, we consider $S$ values of 96, 192, 336 and 720. The results are reported, and the best performance is highlighted in \textbf{bold}.}
\vskip 0.1in
\begin{center}
\begin{footnotesize}
\begin{sc}
\resizebox{\textwidth}{62mm}{
\begin{tabular}{c|c|cc|cc|cc|cc|cc|cc|cc} 
\toprule

\multicolumn{2}{c|}{Models} & \multicolumn{2}{c|}{\textbf{DGCformer}} & \multicolumn{2}{c|}{PatchTST} & \multicolumn{2}{c|}{Crossformer} & \multicolumn{2}{c|}{DLinear} & \multicolumn{2}{c|}{Autoformer} & \multicolumn{2}{c|}{Informer}  & \multicolumn{2}{c}{MTGNN}\\ \midrule

\multicolumn{2}{c|}{Metric} & MSE & MAE & MSE & MAE & MSE & MAE & MSE & MAE & MSE & MAE & MSE & MAE & MSE & MAE  \\ \midrule

\multirow{4}{*}{\rotatebox{90}{ETTh1}} & 96 & \textbf{0.379} & \textbf{0.400} & 0.393 & 0.408 & 0.396 & 0.412 & 0.386 & \textbf{0.400} & 0.449 & 0.459 & 0.865 & 0.713 & 0.522 & 0.490 \\ 

& 192 & \textbf{0.426} & \textbf{0.432} & 0.445 & 0.434 & 0.534 &  0.515 & 0.437 & \textbf{0.432} & 0.500 & 0.482 & 1.008 & 0.792 & 0.542 & 0.536  \\ 

& 336 & \textbf{0.456} & 0.454 & 0.484 & \textbf{0.451} & 0.656 & 0.581 & 0.481 & 0.459 & 0.521 & 0.496 & 1.107 & 0.809 & 0.736 & 0.643 \\ 

& 720 & \textbf{0.458} & \textbf{0.468} & 0.480 & 0.471 & 0.849 & 0.709 & 0.519 & 0.516 & 0.514 & 0.512 & 1.181 & 0.865 & 0.916 & 0.750 \\ \midrule

\multirow{4}{*}{\rotatebox{90}{ETTh2}} & 96 & \textbf{0.278} & \textbf{0.340} & 0.294 & 0.343 & 0.339 & 0.379 & 0.333 & 0.387 & 0.358 & 0.397 & 3.489 & 1.515 & 1.843 & 0.911 \\ 

& 192 & \textbf{0.375} & \textbf{0.388} & 0.377 & 0.393 & 0.415 &  0.425 & 0.477 & 0.476 & 0.456 & 0.452 & 3.755 & 1.525 & 2.439 & 1.325 \\ 

& 336 & \textbf{0.414} & \textbf{0.425} & 0.422 & 0.430 & 0.452 & 0.468 & 0.594 & 0.541 & 0.482 & 0.486 & 4.721 & 1.835 & 2.944 & 1.405 \\ 

& 720 & \textbf{0.415} & \textbf{0.430} & 0.424 & 0.442 & 0.455 & 0.471 & 0.831 & 0.657 & 0.515 & 0.511 & 3.647 & 1.625 & 3.244 & 1.592 \\ \midrule

\multirow{4}{*}{\rotatebox{90}{ETTm1}} & 96 & 0.327 & 0.367 & 0.321 & \textbf{0.360} & \textbf{0.320} & 0.373 & 0.345 & 0.372 & 0.505 & 0.475 & 0.672 & 0.571 & 0.428 & 0.446 \\ 

& 192 & \textbf{0.366} & \textbf{0.382} & 0.379 & 0.388 & 0.386 &  0.401 & 0.380 & 0.389 & 0.553 & 0.496 & 0.795 & 0.669 & 0.457 & 0.469  \\ 

& 336 & \textbf{0.405} & \textbf{0.409} & 0.407 & 0.415 & 0.406 & 0.427 & 0.413 & 0.413 & 0.621 & 0.537 & 1.212 & 0.871 & 0.579 & 0.562 \\ 

& 720 & \textbf{0.462} & \textbf{0.439} & 0.466 & 0.438 & 0.569 & 0.528 & 0.474 & 0.453 & 0.671 & 0.561 & 1.166 & 0.823 & 0.798 & 0.671 \\ \midrule

\multirow{4}{*}{\rotatebox{90}{ETTm2}} & 96 & \textbf{0.176} & \textbf{0.258} & 0.178 & 0.260 & 0.196 & 0.275 & 0.193 & 0.292 & 0.255 & 0.339 & 0.365 & 0.453 & 0.289 & 0.364 \\ 

& 192 & \textbf{0.243} & \textbf{0.304} & 0.249 & 0.307 & 0.248 &  0.317 & 0.284 & 0.362 & 0.281 & 0.340 & 0.533 & 0.563 & 0.456 & 0.492 \\ 

& 336 & \textbf{0.303} & \textbf{0.320} & 0.313 & 0.329 & 0.322 & 0.358 & 0.369 & 0.427 & 0.339 & 0.372 & 1.363 & 0.887 & 1.432 & 0.812 \\ 

& 720 & 0.401 & \textbf{0.397} & \textbf{0.400} & 0.398 & 0.402 & 0.406  & 0.554 & 0.522 & 0.433 & 0.432 & 3.379 & 1.338 & 2.972 & 1.336 \\ \midrule

\multirow{4}{*}{\rotatebox{90}{EXC}} & 96 & \textbf{0.081} & \textbf{0.208} & 0.081 & 0.216 & 0.139 & 0.265 & 0.088 & 0.218 & 0.197 & 0.323 & 0.847 & 0.752 & 0.289 & 0.338 \\ 

& 192 & \textbf{0.167} & \textbf{0.294} & 0.169 & 0.317 & 0.241 &  0.375  & 0.176 & 0.315 & 0.300 & 0.369 & 1.204 & 0.895 & 0.441 & 0.498 \\ 

& 336 & \textbf{0.279} & \textbf{0.337} & 0.305 & 0.416 & 0.392 & 0.468 & 0.313 & 0.427 & 0.509 & 0.524 & 1.672 & 1.036 & 0.934 & 0.741  \\ 

& 720 & \textbf{0.759} & \textbf{0.593} & 0.853 & 0.702 & 1.112 & 0.802 & 0.839 & 0.695 & 1.447 & 0.941 & 2.478 & 1.310  & 1.478 & 1.037 \\ \midrule

\multirow{4}{*}{\rotatebox{90}{WTH}} & 96 & \textbf{0.176} & \textbf{0.221} & 0.180 & 0.221 & 0.185 & 0.248 & 0.196 & 0.255 & 0.266 & 0.336 & 0.300 & 0.384 & 0.189 & 0.252 \\ 

& 192 & 0.225 & \textbf{0.259} & \textbf{0.224} & 0.259 & 0.229 &  0.305 & 0.237 & 0.296 & 0.307 & 0.367 & 0.598 & 0.544 & 0.235 & 0.299 \\ 

& 336 & \textbf{0.277} & \textbf{0.297} & 0.278 & 0.298 & 0.287 & 0.332 & 0.283 & 0.335 & 0.359 & 0.395 & 0.578 & 0.523 & 0.295 & 0.359 \\ 

& 720 & \textbf{0.350} & \textbf{0.346} & 0.350 & 0.346 & 0.356 & 0.398 & 0.345 & 0.381 & 0.419 & 0.428 & 1.059 & 0.741 & 0.440 & 0.481 \\ \midrule

\multirow{4}{*}{\rotatebox{90}{ILI}} & 24 & \textbf{1.366} & \textbf{0.785} & 1.522 & 0.814 & 3.041 & 1.186 & 2.215 & 1.081 & 2.906 & 1.182 & 4.657 & 1.449 & 4.265 & 1.387 \\ 

& 36 & \textbf{1.051} & \textbf{0.679} & 1.430 & 0.834 & 3.406 &  1.232 & 1.963 & 0.963 & 2.585 & 1.038 & 4.650 & 1.463 & 4.777 & 1.496  \\ 

& 48 & \textbf{1.497} & \textbf{0.792} & 1.673 & 0.854 & 3.459 & 1.221 & 2.130 & 1.024 & 3.024 & 1.145 & 5.004 & 1.542 & 5.333 & 1.592\\ 

& 60 & \textbf{1.201} & \textbf{0.719} & 1.529 & 0.862 & 3.640 & 1.305 & 2.368 & 1.096 & 2.761 & 1.114 & 5.071 & 1.543  & 5.070 & 1.552 \\ \midrule

\multirow{4}{*}{\rotatebox{90}{ECL}} & 96 & \textbf{0.175} & \textbf{0.273} & 0.195 & 0.289 & 0.219 & 0.314 & 0.197 & 0.282 & 0.201 & 0.317 & 0.274 & 0.368 & 0.243 & 0.342 \\ 

& 192 & \textbf{0.190} & \textbf{0.280} & 0.199 & 0.289 & 0.231 & 0.322 & 0.196 & 0.285 & 0.222 & 0.334 & 0.296 & 0.386 & 0.298 & 0.364    \\ 

& 336 & \textbf{0.210} & \textbf{0.300} & 0.215 & 0.305 & 0.246 & 0.337 & 0.209 & 0.301 & 0.231 & 0.338 & 0.300 & 0.394 & 0.368 & 0.396\\ 

& 720 & 0.259 & 0.344 & 0.256 & 0.337 & 0.280 & 0.363 & \textbf{0.245} & \textbf{0.333} & 0.254 & 0.361 & 0.373 & 0.439  & 0.422 & 0.410 \\ \midrule
\multicolumn{2}{c|}{$1^{\texttt{st}}$ Count} & \multicolumn{2}{c|}{57} & \multicolumn{2}{c|}{4} & \multicolumn{2}{c|}{1} & \multicolumn{2}{c|}{4} & \multicolumn{2}{c|}{0} & \multicolumn{2}{c|}{0} & \multicolumn{2}{c}{0}\\ \bottomrule
\end{tabular}}
\end{sc}
\end{footnotesize}
\end{center}
\label{fullresults}
\end{table*}

\textbf{Baselines.} Since our method is also transformer forecasting structure, we choose some transformer-based method as baselines. \textbf{Informer} \cite{zhou2021informer}, \textbf{Autoformer} \cite{wu2021autoformer}, \textbf{PatchTST} \cite{nie2022time}, \textbf{Crossformer} \cite{zhang2022crossformer}. Also we choose \textbf{DLinear} \cite{zeng2023transformers} and \textbf{MTGNN} \citep{wu2020Connecting} as the comparison. The details of these papers' introduction are shown in Appendix \ref{app:baseline}. For the criterion of justifying models, we choose MSE and Mean Absolute Error (MAE) as our criterion. 

\textbf{Implementation details.} The configuration of the look-back window for our datasets is meticulously chosen based on prior research. Specifically, for the illness data, we set the look-back window length to 104, while for the other datasets, a length of 96 is employed. This selection is informed by the studies conducted in \citet{zhang2023sageformer,yu2023dsformer}.

The train/validation/test sets are subjected to zero-mean normalization, utilizing the mean and standard deviation of the training set. This normalization process adheres to the ratios detailed in Appendix \ref{app:Dataset}. Regarding the prediction length, we generally opt for values of 96, 192, 336, and 720 across datasets. However, for the illness data, we adapt the prediction lengths to 24, 36, 48, and 60, considering its relatively smaller dataset size.

It is noteworthy that PatchTST \citep{nie2022time} and Crossformer \citep{zhang2022crossformer}, known for their superior performance across benchmarks, serve as our primary baseline models for comparative analysis. Comprehensive details about the baseline models, datasets, and implementation specifics can be found in Appendix \ref{Hyperparameters}.

\subsection{Main Results}

Obviously, a lower MSE or MAE value corresponds to greater precision in prediction outcomes. As indicated in Table \ref{fullresults}, our proposed DGCformer method outperforms alternative approaches in the majority of the datasets. Specifically, it demonstrates superior performance in 57 of the total 64 top-1 cases when considering the complete results. The most significant results are highlighted in \textbf{bold}.
\subsection{Ablation Study}
\label{Ablation}

In our methodology, we advocate for the use of deep graph clustering to categorize variables. A comparative analysis of results with or without the GCL and RFL design reveals the significant impact of these elements in improving forecasting accuracy. Moreover, it is imperative to benchmark our method against traditional clustering techniques. Therefore, we execute ablation studies on the ETTh1 dataset from three unique vantage points: (1) w/o GCL module: involves removing the GCL module; (2) w/o RFL module, involves removing the RFL module; and (3) Covering replacing components with other fundamental clustering methods, such as DTW clustering. These ablation experiments are designed to assess the distinct contributions and significance of each component in our proposed framework.

In our analysis of the results presented in Tables \ref{tab:wogl}, \ref{tab:woed}, and \ref{tab:othercls}, several key findings are discerned. Firstly, models incorporating GCL module exhibit superior performance compared to those lacking this which emphasizes the significance of GCL module and the pivotal role of graph-based supervision in enhancing model effectiveness. Secondly, it is evident that the absence of RFL module is disadvantageous. Thirdly, in a comparative evaluation against other clustering models, we utilized the Dynamic Time Warping (DTW) cluster, a commonly used baseline in time series analysis. Our method demonstrably surpasses the DTW cluster in performance, affirming its effectiveness.

\begin{figure}[t]
    \centering
    \includegraphics[width=0.45\textwidth]{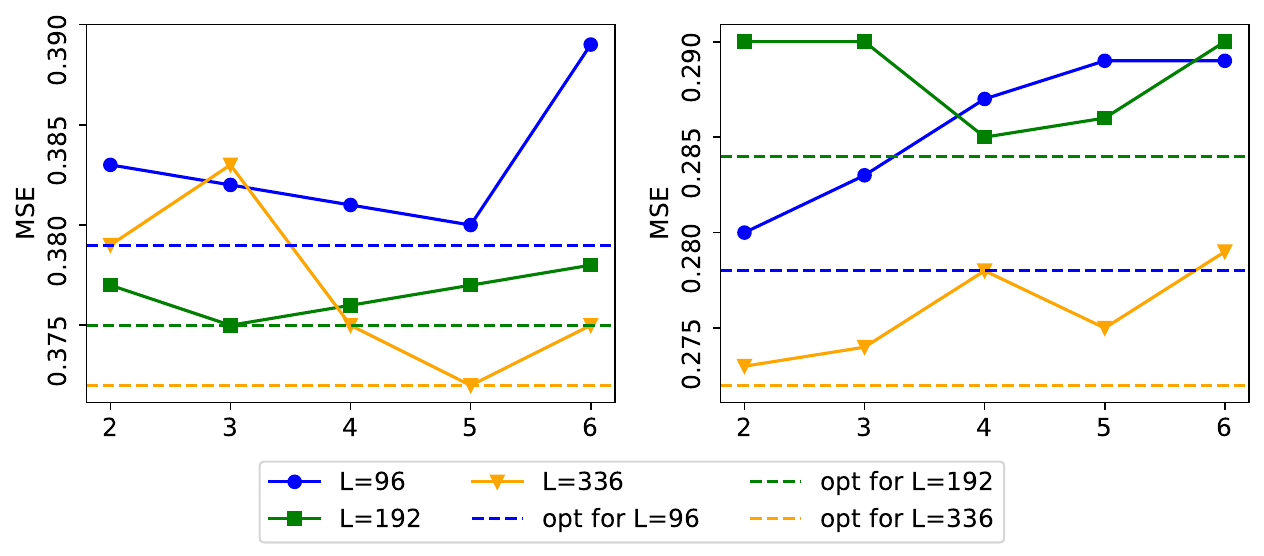}
    \caption{The influence of varying hyperparameters on the ETTh1 and ETTh2 datasets is illustrated in Figures (a) and (b), respectively. The horizontal axis delineates the number of clusters, $n$, utilized within the Graph Clustering Layer (GCL) module, whereas the vertical axis quantifies the Mean Squared Error (MSE) derived from distinct experimental setups. Each trajectory depicted in these figures correlates with a unique look-back window length $L$, highlighting the variable impact of this parameter on the outcomes.}
    \label{fig:hyper}
\end{figure}

\begin{table}[htbp]
    \centering
    \caption{The impact of GCL module on ETTh1 dataset with varying look-back window lengths, ranging from 96 to 720.}
    \vskip 0.15in
    \begin{center}
    \begin{small}
    \begin{sc}
    \begin{tabular}{c|cc|cc}
    \toprule
     GCL &  \multicolumn{2}{c|}{\Checkmark} & \multicolumn{2}{c}{\XSolidBrush} \\
     \midrule
     Metric & MSE & MAE & MSE & MAE \\
     \midrule
     96    & \textbf{0.380} & \textbf{0.400} & 0.381 & 0.406 \\
     192    & \textbf{0.426} & \textbf{0.432} & 0.433 & 0.434 \\
     336    & \textbf{0.456} & 0.454 & 0.478 & \textbf{0.452} \\
     720    & \textbf{0.458} & 0.468 & 0.463 & \textbf{0.466} \\
     \bottomrule
    \end{tabular}
    \end{sc}
    \end{small}
    \end{center}
    \label{tab:wogl}
\end{table}

\begin{table}[htbp]
    \centering
    \caption{The impact of RFL module on ETTh1 dataset with varying look-back window lengths, ranging from 96 to 720.}
    \vskip 0.15in
    \begin{center}
    \begin{small}
    \begin{sc}
    \begin{tabular}{c|cc|cc}
    \toprule
     RFL &  \multicolumn{2}{c|}{\Checkmark} & \multicolumn{2}{c}{\XSolidBrush} \\
     \midrule
     Metric & MSE & MAE & MSE & MAE \\
     \midrule
     96    & \textbf{0.380} & \textbf{0.400} & 0.382 & 0.403 \\
     192    & \textbf{0.426} & \textbf{0.432} & 0.435 & 0.432 \\
     336    & \textbf{0.456} & 0.454 & 0.474 & \textbf{0.450} \\
     720    & \textbf{0.458} & \textbf{0.468} & 0.465 & 0.469 \\
     \bottomrule
    \end{tabular}
    \end{sc}
    \end{small}
    \end{center}
    \label{tab:woed}
\end{table}

\subsection{Hyperparameter Analysis}

In our hyperparameter analysis, we conduct a comprehensive examination to discern the effects of crucial factors on our model's effectiveness. This investigation enables us to evaluate the sensitivity of our method to variations in hyperparameters and to identify the configurations that yield optimal performance. Among these factors, the selection of the appropriate number of clusters $n$, and the determination of the look-back window length emerge as significant contributors to our experimental outcomes.

The influence of cluster numbers on model performance is systematically illustrated in Figure \ref{fig:hyper}, where we conduct targeted experiments on the ETTh1 and ETTh2 datasets by varying the number of clusters. These experiments reveal distinct variations in performance across different cluster configurations. Notably, we introduce an adaptive strategy that selects an optimal cluster number for each batch dynamically, as opposed to a static approach applied uniformly across all batches. As demonstrated in Figure \ref{fig:hyper}, this adaptive method—represented by a distinct line—consistently surpasses the performance of a fixed cluster count in each batch, leading us to incorporate this optimal setting into our experimental framework.

Moreover, the dimension of the look-back window exerts a significant effect on the predictive outcomes. Extending the historical series length tends to enhance prediction accuracy, as it encapsulates a more extensive array of relational data between temporal intervals. Nevertheless, the relationship between window length and accuracy is not unequivocally linear, which may be attributed to potential overfitting with excessively long sequences. To ensure equitable comparison and methodological consistency, we standardize the look-back window length at 96 time steps for our primary analysis, balancing the benefits of comprehensive historical insight against the risks of overfitting.

\begin{table}[t]
    \centering
    \caption{Compared to DTW clustering method on ETTh1 dataset, with varying look-back window lengths, ranging from 96 to 720.}
    \vskip 0.15in
    \begin{center}
    \begin{small}
    \begin{sc}
    \begin{tabular}{c|cc|cc}
    \toprule
      &  \multicolumn{2}{c|}{\textbf{DGCformer}} & \multicolumn{2}{c}{dtw}  \\
     \midrule
     Metric & MSE & MAE & MSE & MAE \\
     \midrule
     96    & \textbf{0.380} & \textbf{0.400} & 0.388 & 0.409 \\
     192    & \textbf{0.426} & \textbf{0.432} & 0.437 & 0.433 \\
     336    & \textbf{0.456} & \textbf{0.454} & 0.479 & 0.455 \\
     720    & \textbf{0.458} & \textbf{0.468} & 0.473 & 0.471 \\
     \bottomrule
    \end{tabular}
    \end{sc}
    \end{small}
    \end{center}
    \label{tab:othercls}
    \vspace{-2.4em}
\end{table}
\section{Conclusion}
This paper presents DGCformer, a transformer-based model merging clustering and forecasting. It clusters variables, follows by applying a former-latter masked self-attention mechanism for within-cluster interactions, and uses a linear layer for predictions. The model incorporates CD and CI to improve predictive accuracy, as demonstrated in standard dataset experiments. In the future work, we aim to develop an integrated, end-to-end model focusing on adaptive clustering and reducing the computational cost.
\section*{Impact Statements}

This paper presents work whose goal is to advance the field of multivariate time series forecasting by using cluster and transformer prediction. There could be many potential positive societal consequences of our work, none of which we feel must be specifically highlighted here.

\bibliography{Master}
\bibliographystyle{icml2024}

\appendix
\onecolumn
\section{Benchmarking Datasets}
\label{app:Dataset}

For the dataset, we conduct experiments on 8 real-world datasets and the details are as followings:
\begin{enumerate}
    \item The Electricity Transformers Temperature (ETT) datasets are produced from two different parts of the same province in China. The dimension of variables is 7 without data. For the details, the variables are HUFL(High UseFul Load), HULL (High UseLess Load), MUFL (Middle UseFul Load), MULL (Middle UseLess Load), LUFL (Low UseFul Load), LULL (Low UseLess Load) and OT (Oil Temperature). ETTh1 and ETTh2 represent the hourly equivalents respectively, which contains 17420 time steps. While ETTm1 and ETTm2 yield a total of 69680 time steps for every 15 minutes.
    \item The Electricity Consumption Load (ECL) dataset records the power usage of 321 customers. The data dimension is (26304,321). And also it performs strongly seasonality.
    \item The Exchange dataset (EXC) collects the daily exchange rates for eight countries (Austrilia, British, Canada, Switherland, China, Japan, New Zealand and Singapore) from 1990 to 2016. The dimension is (7588,8).
    \item The Weather (WTH) dataset includes 21 weather indicators, such as air temperature, humidity, etc. Its data is recorded every 10 minutes in 2020. The dimension is (52696,21).
    \item The Influenza-Like Illness (ILI) dataset describes the ratio of patients with influenza illness to the number of patients. It includes weekly data from the U.S. Centers for Disease Control and Prevention from 2002 to 2021.
\end{enumerate}

And the details of ratio of train/val/test are as following table which are same to \citep{wu2022timesnet}:
\begin{table}[htbp]
    \centering
    \caption{Summarized feature details of eight datasets.}
    \vskip 0.15in
    \begin{center}
    \begin{small}
    \begin{sc}
    \begin{tabular}{c|c|c|c|c|c}
    
    \toprule
     Dataset &  Dimension & Prediction Length & Size (train/val/test) & Frequency & Description \\
     \toprule
     ETTh1    & 7 & $\{96,192,336,512\}$ & 8545/2881/2881 & Hourly & Electricity \\
     \midrule
     ETTh2    & 7 & $\{96,192,336,512\}$ & 8545/2881/2881 & Hourly & Electricity \\
     \midrule
     ETTm1     & 7 & $\{96,192,336,512\}$ & 34465/11521/11521 & 15minutes & Electricity \\
     \midrule
     ETTm2    & 7 & $\{96,192,336,512\}$ & 34465/11521/11521 & 15minutes & Electricity \\
     \midrule
     EXC & 8 & $\{96,192,336,512\}$ & 5120/665/1422 & Daily & Exchange Rate \\
     \midrule
     WTH    & 21 & $\{96,192,336,512\}$ & 36972/5271/10540 & 10minutes & Weather \\
     \midrule
     ILI    & 7 & $\{24,36,48,60\}$ & 617/74/170 & Weekly & Illness \\
     \midrule
     ECL    & 321 & $\{96,192,336,512\}$ & 18317/2633/5261 & Hourly & Electricity \\
     \bottomrule
    \end{tabular}
    \end{sc}
    \end{small}
    \end{center}
    \label{table:results}
\end{table}

\section{Baseline Methods}
\label{app:baseline}
We briefly describe the following baselines:
\begin{enumerate}
    \item \textbf{Informer} \citep{zhou2021informer} is a transformer-based model with ProbSparse self-attention mechanism, which reduce the time complexity from $\mathcal{O}(L^2)$ to $\mathcal{O}(LlogL)$ and memory usage. The source of code is available at {\url{https://github.com/zhouhaoyi/Informer2020}}.
    \item \textbf{Autoformer} \citep{wu2021autoformer} is also a transformer-based model with series decomposition. This model decomposes the original times series into seasonal and trend. The source is available at {\url{https://github.com/thuml/Autoformer}}.
    \item \textbf{Dlinear} \citep{zeng2023transformers} proposes a simple method that even only with linear structure, it can also outperform transformer-based models. The source of code is available at {\url{https://github.com/cure-lab/LTSF-Linear}}.
    \item \textbf{Crossformer} \citep{zhang2022crossformer} proposes a transformer-based method which considers the temporal dependency and variables dependency. Also this paper gives Dimension-Segment-Wise embedding and Two-Stage Attention structure. The code is available at {\url{https://github.com/Thinklab-SJTU/Crossformer}}.
    \item \textbf{PatchTST} \citep{nie2022time} proposes two components: segmentation of time series and channel-independence. Until now it is the sota model used by us and the code is at {\url{https://github.com/yuqinie98/PatchTST}}.
    \item \textbf{MTGNN} \citep{wu2020Connecting} is a model based on graph network which employs temporal convolution and graph convolution layers to capture the time-variable-dimension dependencies. The code is available at {\url{https://github.com/nnzhan/MTGNN}}.
\end{enumerate}

\section{Details of Experiments.}
\label{Hyperparameters}
For the main experiments, we employ the DGCformer, which incorporates a clustering structure followed by a forecasting module. The clustering model comprises an autoencoder structure and a GCN structure, yielding hidden variables that can be categorized into groups. The optimal number of clusters is determined through a grid search conducted by us. By default, the dimension of the hidden representation is set to 32, as determined by us, and after linear projection, the hidden dimension becomes 10, which is used for clustering and establishing the target distribution. Following the extraction of clusters from individual channels, we adopt traditional normalization and apply patching with positional embeddings as proposed in PatchTST. The patch length is set to 16, with a stride of 8, for the majority of datasets, except for dataset Illness, where the values are set to 24 and 2, respectively.
Regarding the former-latter masked self-attention mechanism, we employ 3 encoder layers with a head number of $H=16$ and a latent space dimension of $D=128$. Dropout with a probability of 0.2 is applied within the structures for all experiments.
Furthermore, we utilize mean squared error (MSE) loss for forecasting and entropy loss for clustering as our chosen loss functions. All models are implemented in PyTorch and trained on a single NVIDIA V100 with 32GB memory.

\section{Visualization}

\subsection{Extra Motivation Visualization}

\begin{figure}[htbp]
\centering
\begin{subfigure}{0.8\textwidth}
\begin{center}
\includegraphics[width = 1.0\textwidth]{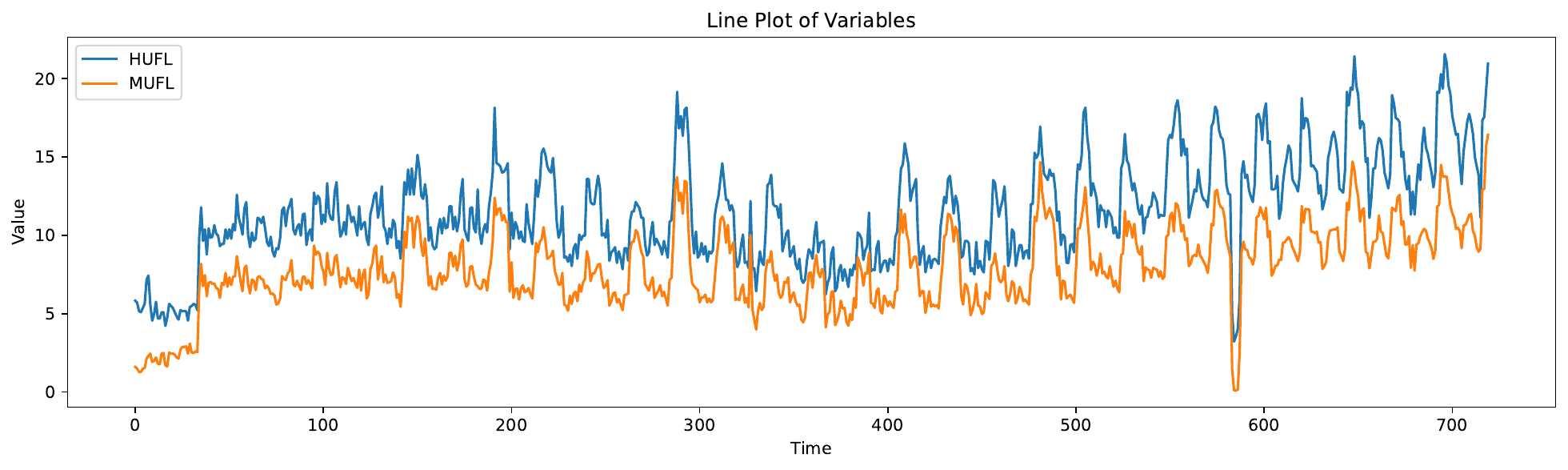}
\caption{}
\end{center}
\end{subfigure}
\begin{subfigure}{0.8\textwidth}
    \begin{center}
        \includegraphics[width = 1.0\textwidth]{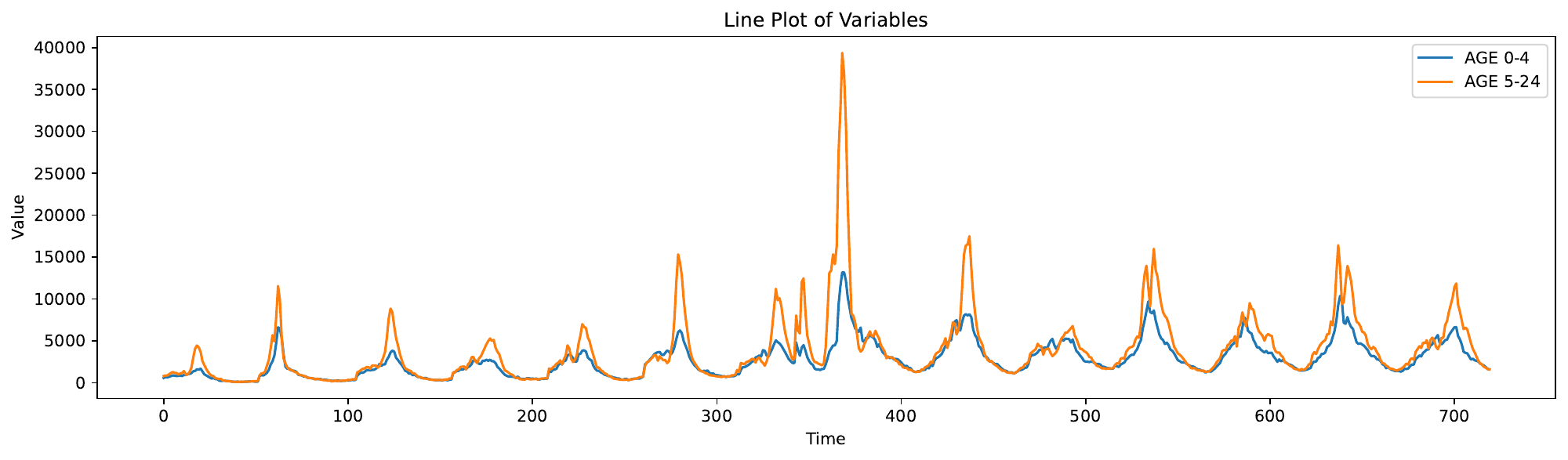}
        \caption{}
    \end{center}
\end{subfigure}
\begin{subfigure}{0.8\textwidth}
    \begin{center}
        \includegraphics[width = 1.0\textwidth]{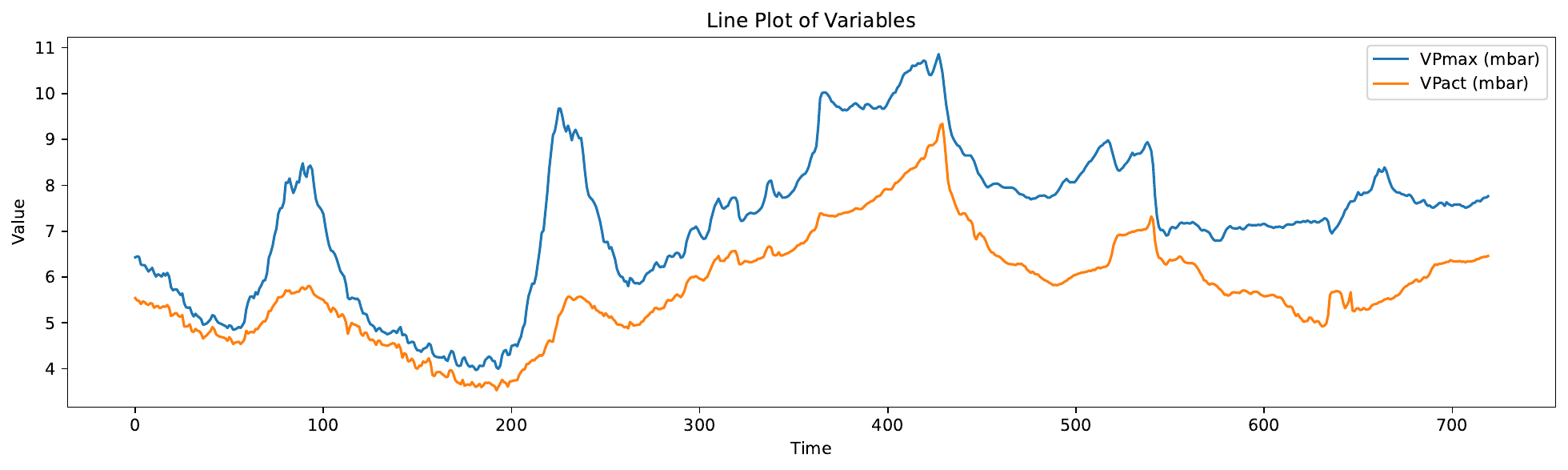}
        \caption{}
    \end{center}
\end{subfigure}
\begin{subfigure}{0.8\textwidth}
    \begin{center}
        \includegraphics[width = 1.0\textwidth]{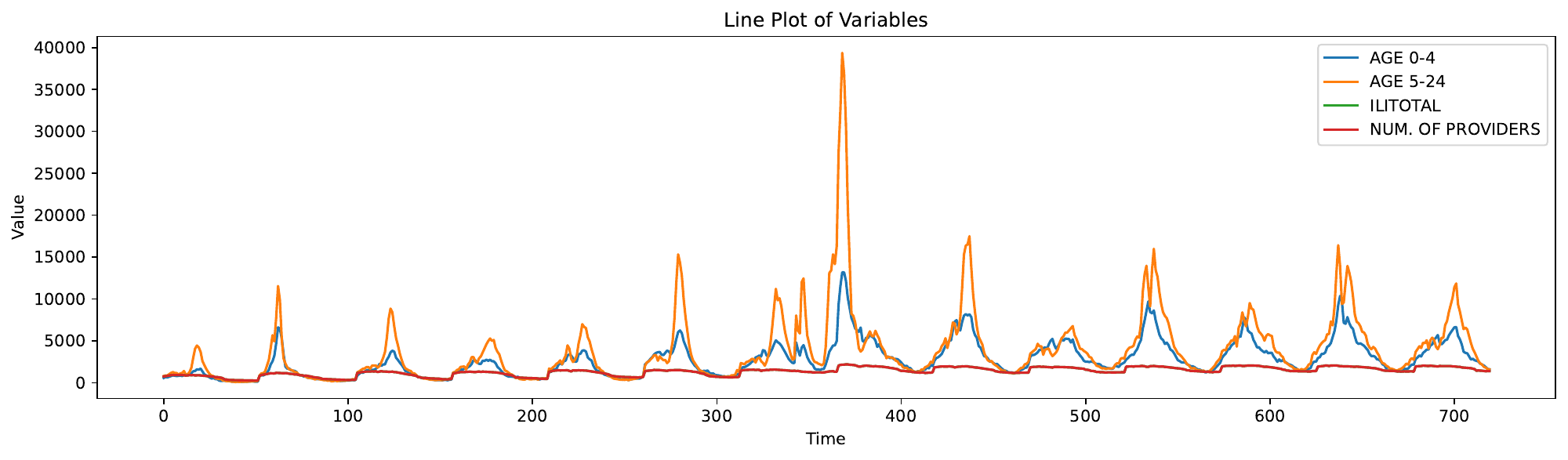}
        \caption{}
    \end{center}
\end{subfigure}
\caption{The figures present the initial 720 time steps of various variables across different datasets. Figure (a) showcases variables HUFL and MUFL from the ETTh1 dataset, while figures (b) and (d) display four variables from the Illness dataset. Figure (c) pertains to the Weather dataset. These illustrations highlight correlations among certain variables, suggesting their potential grouping into clusters for analysis. Notably, the final figure reveals a distinct pattern in the 'Number of Providers' variable, indicating its possible segregation into a separate cluster due to noticeable discrepancies with other variables.}
\label{fig:Visual_plots}
\end{figure}

This section presents additional visualizations explaining our clustering method's application to the entire time series. Analysis of Figure \ref{fig:Visual_plots} identifies correlations among channels in the initial 720 time steps, advocating for channel grouping to overcome CI limitations. This strategy, by treating ungrouped channels as CI, potentially addresses distribution drifts in the datasets. For instance, Figure \ref{fig:Visual_plots}(d) suggests a provisional grouping of the first three variables separately from the fourth. Notably, Figure \ref{fig:Visual_plots}(c) illustrates time-lagged correlations, indicating the inadequacy of simple k-means clustering. Consequently, we propose a graph clustering method, integrating a correlation coefficient matrix, to facilitate more effective groupings. This approach establishes a comprehensive framework for our clustering strategy.

\subsection{Forecasting Visualization}
\begin{figure}[htbp]
\centering
\begin{subfigure}{0.45\textwidth}
\begin{center}
\includegraphics[width = 1.0\textwidth]{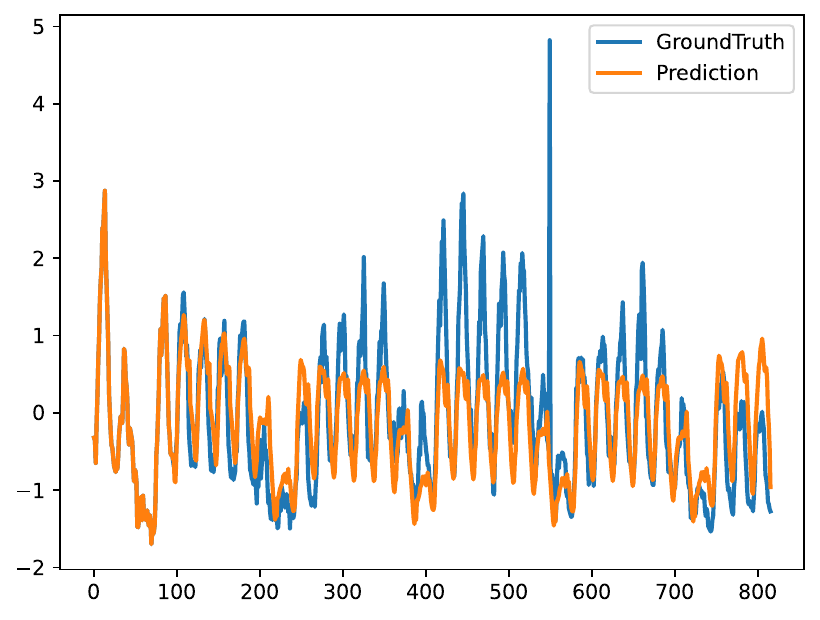}
\caption{}
\end{center}
\end{subfigure}
\begin{subfigure}{0.45\textwidth}
    \begin{center}
        \includegraphics[width = 1.0\textwidth]{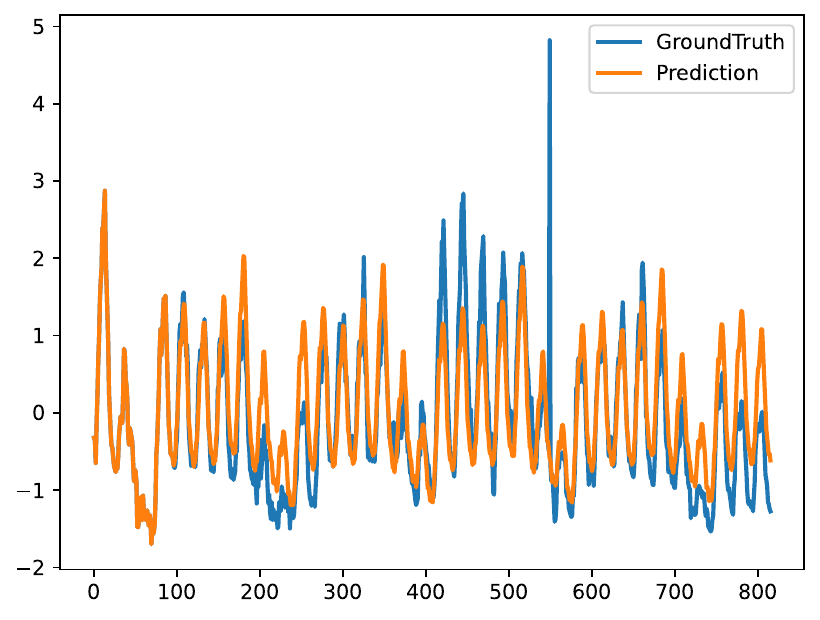}
        \caption{}
    \end{center}
\end{subfigure}
\begin{subfigure}{0.45\textwidth}
    \begin{center}
        \includegraphics[width = 1.0\textwidth]{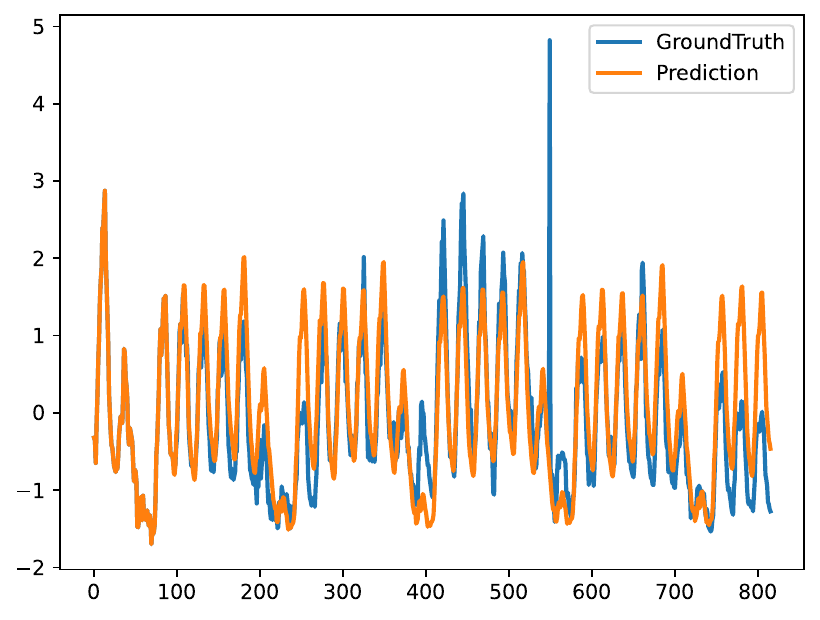}
        \caption{}
    \end{center}
\end{subfigure}
\begin{subfigure}{0.45\textwidth}
    \begin{center}
        \includegraphics[width = 1.0\textwidth]{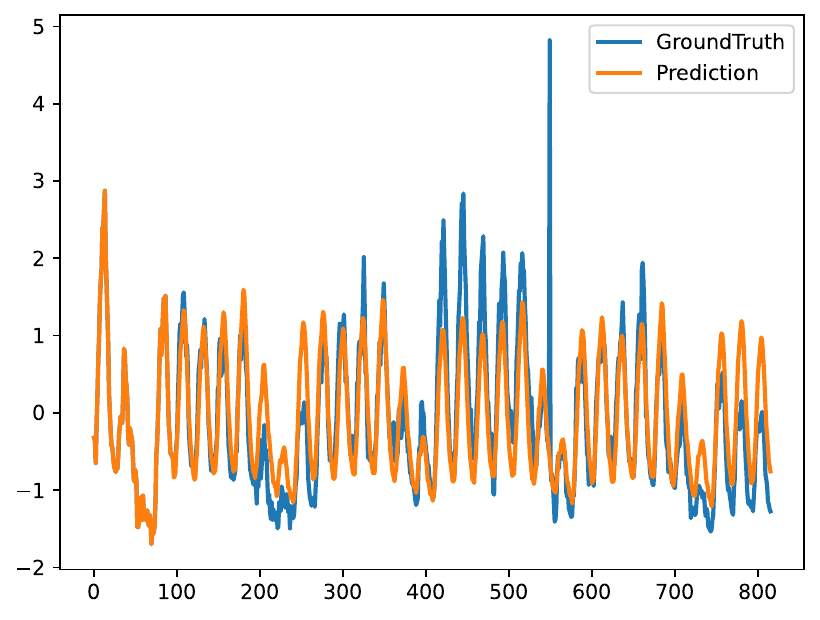}
        \caption{}
    \end{center}
\end{subfigure}
\caption{The visualizations depict results from the Electricity dataset, illustrating the use of a historical length of 96 time steps to forecast the subsequent 720 steps. The figures compare the performance of different methods: (a) Autoformer, (b) DLinear, (c) PatchTST, and (d) DGCformer (our proposed method). Each figure demonstrates the unique forecasting capabilities and results of these methods, showcasing the effectiveness of our DGCformer in comparison to the others.}
\label{fig:Visual_forecasting}
\end{figure}

In addition to basic methods, we present forecasting visualizations for several transformer-based techniques, as shown in Figure \ref{fig:Visual_forecasting}. Figures (a) through (d) display the forecasting results of Autoformer, DLinear, PatchTST, and DGCformer (our method). Notably, the visual comparison reveals that our DGCformer exhibits enhanced performance in specific details of the forecast, as detailed in these figures. This comparative analysis underscores the strengths of DGCformer in relation to these transformer-based forecasting methods.

\end{document}